# Fine-Tuning Over Architectural Complexity: Broad-Coverage PII Detection on PIIBench with DeBERTa

**Pritesh Jha**

priteshjha2711@gmail.com

https://github.com/pritesh-2711/pii-bench

**Abstract.** Personally identifiable information (PII) detection systems are frequently trained within narrow source or domain boundaries, limiting coverage when deployed on heterogeneous text. We study model fine-tuning on a corrected multi-source PIIBench preparation spanning 82 retained entity types across ten source datasets. We evaluate three DeBERTa-based approaches: direct token-classification fine-tuning, a source-conditioned hierarchical model (SC+H), and a three-phase curriculum extension (SC+H+Curr). Against eight published comparator systems on a reproducible 5,000-record held-out subset (test_5k), direct fine-tuned DeBERTa achieves F1 0.6476, while SC+H and the curriculum variant achieve 0.5899 and 0.2772 respectively; the strongest published comparator reaches only 0.1723. Because validation initially favoured SC+H, we perform a final streamed evaluation on the complete 100,002-record held-out split. Direct fine-tuning remains superior, achieving F1 0.6455 versus 0.5894 for SC+H. Entity-level analysis shows that direct fine-tuning wins 54 of 82 fine entity types and all ten coarse groups by support-weighted entity F1, while SC+H retains localised advantages on 28 types. The results indicate that diverse task-specific training data and a simple weighted cross-entropy objective contribute more to broad-coverage PII detection than the tested architectural and curriculum complexity.

---

## 1. Introduction

The detection and redaction of personally identifiable information (PII) is a prerequisite for regulatory compliance under GDPR, HIPAA, and CCPA. Production systems processing user-generated or enterprise text must identify spans corresponding to sensitive entities—names, contact details, financial identifiers, credentials—and protect them before downstream use. Despite the practical importance, no single published model or rule-based engine covers the full taxonomy of PII types that appear across real-world corpora.

PIIBench is a unified benchmark consolidating ten heterogeneous source datasets into a standardised BIO token-classification task. Evaluating eight published systems on a stratified held-out subset, the original paper established that all systems achieve span-level F1 below 0.14, directly quantifying the domain-silo problem: each system covers at most a small slice of the full PII taxonomy it encounters in deployment.

Against this backdrop, the performance gap is stark. **The best published system on our corrected dataset benchmark achieves span-level F1 of 0.1723 (SpanMarker BERT). Our direct fine-tuned DeBERTa reaches F1 0.6476 on the same 5,000-record held-out subset**---an absolute gain of **+0.475 F1** and a **3.76x relative improvement**---and F1 0.6455 on the full 100,002-record test split. Even the weakest of our three trained models (SC+H with curriculum, F1 0.2772) comfortably exceeds every published comparator. Figure 1 illustrates the gap across all eleven evaluated systems.

The original PIIBench paper characterises the problem. The present work proposes and evaluates a solution. We ask whether a DeBERTa model fine-tuned on the corrected multi-source preparation can substantially exceed the published comparators, and whether architectural extensions—source conditioning, hierarchical coarse-to-fine prediction, and sequential curriculum learning—improve on simpler direct fine-tuning.

The corrected dataset preparation addresses processing issues identified after the original release: correct Nemotron span parsing, a revised retained label space of 82 entity types, rebalanced Nemotron representation, and corrected train/validation/test membership. Because this updated preparation changes split membership, we re-evaluate all eight comparators on a new reproducible 5,000-record subset drawn exclusively from the corrected test split.

**Contributions.** This paper makes the following contributions:

- A corrected multi-source PIIBench preparation with updated Nemotron parsing, an 82-type label space, balanced source mixture, and SHA-256-identified evaluation artifacts.
- Direct DeBERTa fine-tuning as a strong task-specific baseline, achieving span-level F1 0.6455 on the 100,002-record full held-out split—a gain of 0.507 F1 over the strongest published comparator on the same corrected benchmark.
- A source-conditioned hierarchical architecture (SC+H) as a controlled ablation, with entity-level analysis identifying the categories where additional structure helps.
- Empirical evidence that sequential source-family curriculum learning degrades broad-coverage detection via catastrophic forgetting of prior source families.
- A complete reproducible evaluation pipeline with SHA-256-identified test artifacts and code paths made publicly available.

## 2. Related Work

### 2.1 Named Entity Recognition Benchmarks

CoNLL-2003 (Sang and De Meulder, 2003) is the canonical NER benchmark, covering four entity types over English and German newswire. OntoNotes 5.0 (Weischedel et al., 2013) extends coverage to 18 types across multiple genres. WikiANN (Pan et al., 2017) provides cross-lingual annotations for 282 languages. MultiNERD (Tedeschi and Navigli, 2022) introduces 15 fine-grained types, and Few-NERD (Ding et al., 2021) provides 66 types for episodic learning. None of these benchmarks is designed for PII detection: their taxonomies include non-sensitive types and omit critical PII categories such as financial identifiers and credentials.

### 2.2 PII-Specific Datasets

The ai4privacy dataset series (ai4privacy, 2023) provides large-scale synthetic PII annotations with 54–63 label classes across multiple languages. The Gretel synthetic finance dataset (Gretel.ai, 2023) covers 29 PII types across 100 financial document formats. NVIDIA's Nemotron-PII dataset (NVIDIA, 2023) covers 55+ entity types across 50 industries. Each defines an independent label vocabulary, preventing direct cross-dataset evaluation without normalisation.

### 2.3 Financial NER

FiNER-139 (Loukas et al., 2022) annotates 1.1 million sentences from SEC 10-K/10-Q filings with 139 XBRL financial disclosure tags. While not a classical PII dataset, it provides dense supervision for financial entities underrepresented in general NER corpora.

### 2.4 PII Detection Systems

Microsoft Presidio (Microsoft, 2023) combines regular-expression patterns with spaCy-backed NLP to identify 18 PII entity types. Transformer-based general NER models—BERT-base NER, XLM-RoBERTa NER, SpanMarker mBERT, SpanMarker BERT—are widely used but trained exclusively on general NER labels. PII-specific models include Piiranha DeBERTa, fine-tuned on ai4privacy-400k, and XtremeDistil FiNER, fine-tuned on FiNER-139 XBRL tags. All are siloed to their training domain.

### 2.5 Multi-Domain Adaptation and Curriculum Learning

Sequential curricula can improve data-efficiency when source families have compatible structures, but are prone to catastrophic forgetting (McCloskey and Cohen, 1989) when later phases specialise away from earlier distributions. Our experiments provide direct evidence of this failure mode in the PII setting.

### 2.6 PIIBench

Jha (2026) presented the original PIIBench corpus: 2,369,883 annotated sequences across 48 canonical entity types from ten source datasets, with a normalisation pipeline mapping 80+ source-specific label variants to a standardised BIO scheme. Eight published systems evaluated on the 1,398-record stratified test subset all achieved F1 below 0.14 (best: Presidio at 0.1385). The present work extends PIIBench with a corrected data preparation and trained model evaluation.

## 3. Corrected Data Preparation

### 3.1 Motivation for Correction

Post-release exploratory analysis of the original PIIBench preparation identified two processing issues: (1) NVIDIA Nemotron PII spans were not parsed correctly from their XML-tagged text format, producing incorrect BIO annotations; (2) Nemotron records were not proportionally balanced, resulting in under-representation. These issues affected train/validation/test membership and motivated a corrected preparation.

### 3.2 Processing Pipeline

The corrected pipeline applies the following steps in order:

1. Consolidate all retained source datasets to canonical BIO annotations.
2. Re-parse NVIDIA Nemotron PII spans from XML-tagged text and retain only PII-bearing records.
3. Balance Nemotron to approximately 10% of the prepared corpus through stratified sampling.
4. Cap `finer_139` at 150,000 records to prevent financial filings from dominating the mixture.
5. Remove rare entity types with fewer than a configured threshold of B-tag mentions.
6. Produce source-stratified train, validation, and test JSONL splits.p

### 3.3 Prepared Splits

Table 1 reports the resulting split sizes. Training data totals 799,948 records. The 100,002-record test split is the primary evaluation artifact. The test_5k subset of 5,000 records is used for multi-system comparative evaluation.

| Split | Records | Use |
|---|---|---|
| Train | 799,948 | Model optimisation |
| Validation | 99,990 | Held-out validation population |
| `val_1p` | 994 | Fast intra-training model selection |
| Test | 100,002 | Final full held-out evaluation |
| `test_5k` | 5,000 | Multi-system comparative evaluation |

*Table 1: Corrected PIIBench splits used in this work.*

### 3.4 Label Space and Class Imbalance

After rare-label filtering, 82 entity types are retained. With BIO prefixes and the outside label O, the fine label space contains 165 classes. The hierarchical coarse taxonomy groups these into ten categories plus O, yielding 21 coarse BIO labels. Table 2 summarises the token-level class distribution.

| Item | Value |
|---|---|
| Retained entity types | 82 |
| Fine BIO labels (incl. O) | 165 |
| Coarse BIO labels (incl. O) | 21 |
| Training supervised tokens | 52,846,281 |
| O tokens | 44,016,269 (83.3%) |
| Non-O tokens | 8,830,012 (16.7%) |

*Table 2: Label space and token imbalance in the training split.*

The 83% outside-token majority causes training collapse under uniform cross-entropy: all models trained without reweighting produced near-zero precision, recall, and F1. All reported trained models therefore use a weighted loss with w(O) = 0.1 and w(entity) = 1.0.

### 3.5 test_5k Evaluation Subset

The `test_5k` artifact is selected from the corrected test split by source-stratified proportional sampling with largest-remainder allocation and random seed 42. The subset contains 222 BIO continuation labels that seqeval interprets as new entity spans. Since every model is evaluated on the identical file with the identical metric, this does not affect relative rankings, but we disclose it as an annotation caveat.

| Item | Value |
|---|---|
| Records | 5,000 |
| Gold spans (seqeval) | 15,548 |
| Represented entity types | 82 |
| Represented sources | 10 |
| SHA-256 | `46238c6cc12526d4d29e9282e96d2936...` |

*Table 3: test_5k evaluation subset properties.*

## 4. Methods

We train three DeBERTa-based models of increasing architectural complexity, using `microsoft/deberta-v3-base` (He et al., 2023) as the shared backbone.

### 4.1 Model A: Direct Fine-Tuned DeBERTa

The simplest model fine-tunes deberta-v3-base as a standard BIO token classifier with weighted cross-entropy. No source metadata, auxiliary tasks, or curriculum phases are used. A linear classification head maps each token's contextualised representation to a distribution over 165 fine BIO labels. This model is released on HuggingFace Hub as piibench-deberta-base.

| Parameter | Value |
| --- | --- |
| Max sequence length | 256 |
| Batch size per device | 6 |
| Gradient accumulation | 10 |
| Effective batch size | 60 |
| Learning rate | 2e-5 |
| O / entity loss weights | 0.1 / 1.0 |
| Validation subset | `val_1p`, span-level F1 |
| Best validation step | 24,000 |

*Table 4: Direct fine-tuning hyperparameters.*

### 4.2 Model B: Source-Conditioned Hierarchical DeBERTa (SC+H)

Released on HuggingFace Hub as `piibench-deberta-sch`.

The first proposed extension adds two components.

A learned dataset-source token is prepended to each input sequence (e.g., `[SRC=ai4privacy]`), conditioning the encoder on data provenance. At inference, `[SRC=general]` is used for out-of-distribution inputs.

A coarse-to-fine hierarchical classifier is applied to the DeBERTa hidden states. The coarse head predicts over ten entity groups; its output probabilities are concatenated with the encoder hidden states before the fine BIO head.

Training optimises a combined loss: L = L_fine + 0.3 × L_coarse.

The ten coarse groups are: PERSON_GROUP, CONTACT, FINANCIAL_ID, TEMPORAL, CREDENTIAL, NETWORK, ORG_ROLE, LOCATION, MISC, and FINANCIAL_NER.

| Parameter | Value |
| --- | --- |
| Max sequence length | 256 |
| Effective batch size | 60 |
| Learning rate | 2e-5 |
| Coarse loss weight | 0.3 |
| O / entity loss weights | 0.1 / 1.0 |
| Gradient checkpointing | Enabled |
| Best validation step | 37,000 |

*Table 5: SC+H hyperparameters.*

### 4.3 Model C: SC+H with Curriculum (SC+H+Curr)

The most complex model training uses the Source Conditioned Hierarchical architecture and applies three sequential one-epoch training phases ordered by source family.

Table 6 reports the val_1p F1 at the end of each phase.

| Phase | Source Family | val_1p F1 |
|---|---|---|
| 1 | General NER | 0.131 |
| 2 | Synthetic PII | 0.430 |
| 3 | Financial PII (`gretel_finance`, `finer_139`) | 0.305 |

*Table 6: Curriculum training phases and val_1p F1 at phase completion.*

F1 improves substantially through Phase 2 but drops following Phase 3, providing direct evidence of catastrophic forgetting: the financial-domain specialisation overwrites the broader PII representations acquired in Phase 2.

## 5. Experimental Setup

### 5.1 Training Infrastructure

| Field | Value |
|---|---|
| GPU | 1 × NVIDIA L4 (24 GB) |
| Machine type | `g2-standard-12` |
| vCPU / memory | 12 vCPUs / 48 GB |
| Zone | `asia-southeast1-c` |
| Mixed precision | BF16 |
| Max sequence length | 256 tokens |

*Table 7: GCP training infrastructure.*

Direct DeBERTa training ran for approximately 11.8 hours before best-model export. SC+H completed 39,999 optimiser steps in approximately 17 hours 54 minutes. Full-test inference was run locally on an NVIDIA GeForce RTX 4070 (8 GB VRAM) using a memory-efficient streaming evaluator.

### 5.2 Evaluation Protocol

All precision, recall, and F1 values use span-level seqeval scoring, requiring exact match of both entity span boundaries and entity-type label. No partial credit is awarded.

The `test_5k` evaluation places all eleven systems on the identical test file. The full 100,002-record evaluation is reserved for the head-to-head competition between Direct DeBERTa and SC+H. To avoid prediction-accumulation memory issues, full-test inference uses a custom streaming evaluator: records are read in chunks of 5,000, inference is batched in smaller minibatches, and seqeval metrics are accumulated incrementally without retaining record-level predictions in memory.

## 6. Results

### 6.1 Historical Context: Original PIIBench Comparators

Table 8 reports the eight comparator systems on the original PIIBench 1,398-record test subset (Jha, 2026). These scores motivated the present experiments; the corrected test_5k scores in Table 9 are the appropriate comparators for the newly trained models.

| System | Type | F1 | P | R |
|---|---|---|---|---|
| Microsoft Presidio | Rule-based | 0.1385 | 0.1522 | 0.1271 |
| `spaCy en_core_web_lg` | General NER | 0.0873 | 0.0636 | 0.1395 |
| SpanMarker BERT | General NER | 0.0877 | 0.1723 | 0.0588 |
| SpanMarker mBERT | General NER | 0.0823 | 0.1477 | 0.0570 |
| BERT-base NER | General NER | 0.0657 | 0.0791 | 0.0562 |
| XLM-RoBERTa NER | General NER | 0.0494 | 0.0508 | 0.0482 |
| Piiranha DeBERTa | PII-specific | 0.0463 | 0.0604 | 0.0375 |
| XtremeDistil FiNER | Financial NER | 0.0413 | 0.6990 | 0.0213 |

*Table 8: Original PIIBench comparators on the 1,398-record test subset (Jha, 2026).*

## 6.2 Corrected test_5k Complete Comparison

Table 9 reports all eleven systems on the corrected test_5k subset. Direct DeBERTa achieves F1 0.6476, an absolute gain of 0.475 over the strongest published comparator (SpanMarker BERT at 0.1723). The curriculum variant at 0.2772 still outperforms every published comparator by at least 0.105 F1, demonstrating that even imperfect multi-source fine-tuning substantially widens the gap over domain-siloed systems.

| System | Type | F1 | P | R |
|---|---|---|---|---|
| **Direct DeBERTa** | Fine-tuned | **0.6476** | **0.6300** | **0.6662** |
| SC+H | Proposed | 0.5899 | 0.5565 | 0.6274 |
| SC+H+Curr | Proposed+Curr | 0.2772 | 0.3491 | 0.2299 |
| SpanMarker BERT | General NER | 0.1723 | 0.4266 | 0.1080 |
| SpanMarker mBERT | General NER | 0.1622 | 0.3659 | 0.1042 |
| Microsoft Presidio | Rule-based | 0.1421 | 0.1748 | 0.1197 |
| BERT-base NER | General NER | 0.1321 | 0.1834 | 0.1032 |
| XLM-RoBERTa NER | General NER | 0.1153 | 0.1373 | 0.0993 |
| Piiranha DeBERTa | PII-specific | 0.0872 | 0.1204 | 0.0684 |
| `spaCy en_core_web_lg` | General NER | 0.0853 | 0.0677 | 0.1152 |
| XtremeDistil FiNER | Financial NER | 0.0191 | 0.6089 | 0.0097 |

*Table 9: All systems on the corrected test_5k subset. Bold = best per metric.*

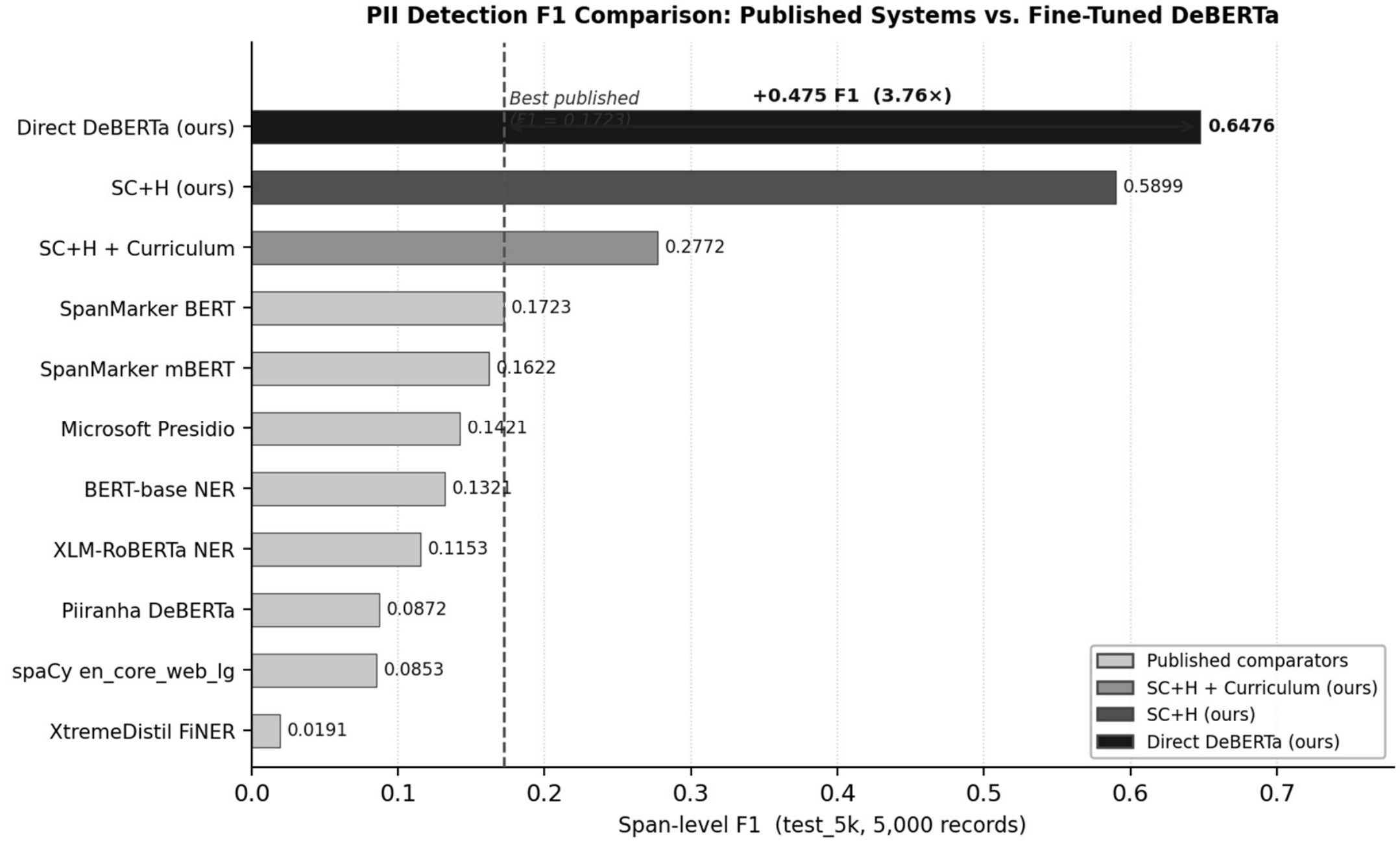

**Figure 1:** Span-level F1 on test_5k (5,000 records) for all eleven evaluated systems. Light grey = published comparators; dark bars = fine-tuned DeBERTa models. Dashed line marks the best published system (SpanMarker BERT, F1 = 0.1723). Direct DeBERTa achieves F1 0.6476 — a **3.76x improvement and +0.475 absolute F1 gain** over the best published comparator.

### 6.3 Validation–Test Disagreement and Final Full-Test Evaluation

The fast `val_1p` subset showed a marginal advantage for SC+H (F1 0.661 vs. 0.645 for Direct DeBERTa). The `test_5k` result reversed this, favouring Direct DeBERTa by 0.058. To resolve the disagreement, both models were evaluated on the complete 100,002-record test split.

| Evaluation | Direct F1 | SC+H F1 | Gap | Winner |
|---|---|---|---|---|
| `val_1p` | 0.6445 | 0.6612 | +0.017 SC+H | SC+H |
| `test_5k` | 0.6476 | 0.5899 | +0.058 Direct | Direct |

*Table 10: Validation–test disagreement between Direct DeBERTa and SC+H.*

| Model | Records | F1 | P | R |
|---|---|---|---|---|
| **Direct DeBERTa** | 100,002 | **0.6455** | **0.6277** | **0.6645** |
| SC+H | 100,002 | 0.5894 | 0.5560 | 0.6270 |
| *Direct − SC+H* | — | +0.0561 | +0.0717 | +0.0375 |

*Table 11: Final full-test evaluation (100,002 records, 308,421 gold spans). Bold = best per metric.*

Direct DeBERTa is superior on all three metrics across the full held-out distribution. This is the primary result: the simplest fine-tuning approach provides the strongest broad-coverage PII detection.

# 7. Entity-Level Analysis

## 7.1 Overall Type-Level Winner Counts

Of the 82 retained fine entity types, Direct DeBERTa achieves higher F1 on 54 types and SC+H on 28 types, with no ties. Direct DeBERTa wins all ten coarse groups by support-weighted mean fine-entity F1.

## 7.2 Coarse Group Summary

Table 12 reports support-weighted mean F1 per coarse group on the full test split. Direct DeBERTa wins every group. The largest advantage is on FINANCIAL_ID, where direct fine-tuning wins all 11 constituent entity types and holds a 0.146 F1 advantage—despite the curriculum variant having ended with financial-domain specialisation.

| Group | Support | Direct F1 | SC+H F1 | Δ | Wins D/S |
|---|---|---|---|---|---|
| FINANCIAL_NER | 58,821 | 0.3229 | 0.2412 | +0.082 | 1/0 |
| LOCATION | 53,111 | 0.7151 | 0.6729 | +0.042 | 6/4 |
| PERSON_GROUP | 46,789 | 0.8004 | 0.7515 | +0.049 | 5/1 |
| ORG_ROLE | 30,723 | 0.7422 | 0.7252 | +0.017 | 2/3 |
| TEMPORAL | 30,683 | 0.5923 | 0.5548 | +0.038 | 2/2 |
| NETWORK | 24,406 | 0.6611 | 0.5920 | +0.069 | 5/4 |
| MISC | 23,574 | 0.7318 | 0.6321 | +0.100 | 11/5 |
| CONTACT | 18,437 | 0.7087 | 0.6593 | +0.049 | 2/2 |
| CREDENTIAL | 12,882 | 0.8902 | 0.8611 | +0.029 | 9/7 |
| FINANCIAL_ID | 8,995 | 0.8763 | 0.7305 | +0.146 | 11/0 |

*Table 12: Coarse group results on the full test split. Δ = Direct F1 − SC+H F1.*

## 7.3 Entity Types Most Favouring Direct Fine-Tuning

| Entity | Group | Supp. | Direct | SC+H | Δ |
|---|---|---|---|---|---|
| CRYPTO_ADDRESS | MISC | 1,569 | 0.8641 | 0.0012 | +0.863 |
| VEHICLE | MISC | 591 | 0.9787 | 0.4689 | +0.510 |
| IBAN | FINANCIAL_ID | 769 | 0.8695 | 0.4952 | +0.374 |
| ACCOUNT_NUMBER | FINANCIAL_ID | 2,686 | 0.9095 | 0.6505 | +0.259 |
| PHONE | CONTACT | 1,535 | 0.9954 | 0.8213 | +0.174 |
| IP_ADDRESS | NETWORK | 6,178 | 0.5528 | 0.4198 | +0.133 |
| SSN | CREDENTIAL | 1,261 | 0.9517 | 0.8261 | +0.126 |
| NAME | PERSON_GROUP | 6,645 | 0.5464 | 0.4355 | +0.111 |
| USERNAME | NETWORK | 8,287 | 0.5285 | 0.4185 | +0.110 |
| FINANCIAL_ENTITY | FINANCIAL_NER | 58,821 | 0.3229 | 0.2412 | +0.082 |

*Table 13: Top 10 entity types by Direct DeBERTa advantage on the full test split.*

### 7.4 Entity Types Most Favouring SC+H

| Entity | Group | Supp. | Direct | SC+H | Δ |
|---|---|---|---|---|---|
| HTTP_COOKIE | NETWORK | 595 | 0.1812 | 0.5752 | +0.394 |
| LOCAL_LATLNG | LOCATION | 55 | 0.3810 | 0.4655 | +0.085 |
| BLOOD_TYPE | MISC | 462 | 0.8265 | 0.8862 | +0.060 |
| DATE_TIME | TEMPORAL | 1,463 | 0.7422 | 0.7844 | +0.042 |
| COUNTY | LOCATION | 611 | 0.8980 | 0.9275 | +0.030 |
| COORDINATE | LOCATION | 857 | 0.8874 | 0.9068 | +0.020 |
| EDUCATION_LEVEL | MISC | 1,137 | 0.7623 | 0.7799 | +0.018 |
| PHONE_NUMBER | CONTACT | 3,079 | 0.8606 | 0.8738 | +0.013 |
| STATE | LOCATION | 1,974 | 0.8291 | 0.8384 | +0.009 |
| COMPANY_NAME | ORG_ROLE | 8,674 | 0.8487 | 0.8560 | +0.007 |

*Table 14: Top 10 entity types by SC+H advantage on the full test split.*

SC+H's largest advantage is for HTTP_COOKIE (+0.394), a contextually complex type where source provenance information may provide a useful disambiguation signal. These SC+H gains are concentrated on lower-support types and are outweighed by the direct model's advantages on high-frequency categories such as FINANCIAL_ENTITY, IP_ADDRESS, USERNAME, ACCOUNT_NUMBER, PHONE, and SSN.

## 8. Discussion

### 8.1 Simplicity Wins on Broad Coverage

The primary result is that a directly fine-tuned DeBERTa model on broad, correctly prepared multi-source PII data substantially outperforms all tested comparators and both architectural extensions. The F1 improvement over the strongest published system (0.475 on test_5k) is an order of magnitude larger than the difference between Direct DeBERTa and SC+H (0.058 on test_5k, 0.056 on full test). The data preparation quality and the weighted loss function account for more of the performance gain than the architectural choices.

### 8.2 Architectural Complexity and Localised Trade-offs

SC+H is not uniformly weaker. Its source-conditioned hierarchical representations produce meaningful F1 gains on 28 of 82 entity types, particularly contextually grounded types such as HTTP_COOKIE and DATE_TIME. A hybrid system that selectively applies hierarchical decoding for certain coarse categories while using direct decoding for others could potentially recover these localised gains without incurring the broad-coverage regression.

### 8.3 Catastrophic Forgetting in Sequential Curriculum

The curriculum variant provides a clear illustration of catastrophic forgetting: after Phase 2's substantial gain (0.430 on val_1p), Phase 3 financial specialisation drops performance to 0.305. Despite ending with financial-domain training, SC+H+Curr is substantially weaker than Direct DeBERTa on FINANCIAL_ID types. The curriculum schedule tested (General NER → Synthetic PII → Financial PII) is one of many possible orderings; the result demonstrates that naive sequential specialisation is harmful but does not exclude all curriculum designs.

### 8.4 Validation–Test Reversal

The val_1p model selection metric marginally preferred SC+H (0.661 vs. 0.645), but the larger test_5k and full test evaluations reversed this by 0.058 and 0.056 respectively. The reversal indicates that the 994-record fast-validation subset carries insufficient statistical resolution to distinguish models in the 0.01–0.02 F1 range.

### 8.5 Operational Implications

For practitioners deploying PII detection on heterogeneous text, the results support a straightforward approach: collect broad, multi-domain annotated data, correct any source-specific parsing issues, apply weighted cross-entropy to handle class imbalance, and fine-tune a strong pretrained encoder directly. Architectural elaboration should be motivated by demonstrated failure modes on specific entity categories rather than assumed a priori.

## 9. Limitations and Future Work

**Dataset preparation change.** The corrected preparation differs from the original PIIBench paper's preparation. Trained models must therefore be compared with the rerun comparators on test_5k rather than directly with the original Table 5 numbers.

**BIO continuation caveat.** The test_5k artifact contains 222 BIO continuation labels that seqeval treats as new spans. All systems are affected equally for relative ranking. A BIO-normalised sensitivity analysis is deferred to future work.

**Single curriculum schedule.** The curriculum experiment tests one specific ordered source-family schedule. The negative result does not preclude all curriculum designs.

**Single training seed.** No repeated-seed training runs were performed. Architecture effects cannot be separated from checkpoint variance without multi-seed replication.

**No statistical confidence intervals.** Bootstrap confidence intervals on full-test predictions are not computed. Formal significance testing is deferred.

**English only; domain coverage gaps.** PIIBench covers general, synthetic PII, and financial domains but excludes medical records, legal contracts, and social media. Extending coverage and adding multilingual evaluation tracks are planned directions.

## 10. Conclusion

We have presented a fine-tuning study on PIIBench, a corrected multi-source PII detection benchmark spanning 82 entity types and ten heterogeneous source datasets. Against eight published comparator systems—rule-based, general NER, PII-specific, and financial NER—a directly fine-tuned DeBERTa model achieves F1 0.6455 on the 100,002-record full held-out split, an absolute gain of more than 0.50 F1 over the strongest published comparator.

A source-conditioned hierarchical extension (SC+H) is competitive but consistently weaker across the full held-out distribution, winning only 28 of 82 entity types. A sequential curriculum variant (SC+H+Curr) degrades substantially due to catastrophic forgetting in its financial-domain final phase.

The principal finding is that broad-coverage PII detection benefits more from correct multi-source data preparation and a simple weighted training objective than from the tested architectural and curriculum complexity. SC+H retains localised advantages on 28 entity types where source conditioning and coarse supervision appear beneficial, warranting further investigation.

The code, evaluation artifacts, and trained model checkpoints are available at https://github.com/pritesh-2711/pii-bench. The dataset is available at https://huggingface.co/datasets/pritesh-2711/pii-bench. Fine-tuned models are released on HuggingFace Hub: `piibench-deberta-base` (direct fine-tune) and `piibench-deberta-sch` (SC+H).

## Appendix A: Complete Fine-Entity Full-Test Comparison

All 82 retained fine entity types on the complete 100,002-record held-out split. Δ = Direct F1 − SC+H F1. D = Direct DeBERTa wins; S = SC+H wins.

| Entity | Group | Supp. | Dir.F1 | SC+H | Δ | W |
|---|---|---|---|---|---|---|
| ACCOUNT_NUMBER | FINANCIAL_ID | 2,686 | 0.9095 | 0.6505 | +0.259 | D |
| BBAN | FINANCIAL_ID | 102 | 0.5769 | 0.5202 | +0.057 | D |
| BIC | FINANCIAL_ID | 231 | 0.9360 | 0.8148 | +0.121 | D |
| CREDIT_CARD | FINANCIAL_ID | 2,305 | 0.9115 | 0.8088 | +0.103 | D |
| CREDIT_CARD_NUMBER | FINANCIAL_ID | 95 | 0.5422 | 0.5287 | +0.013 | D |
| CREDIT_DEBIT_CARD | FINANCIAL_ID | 899 | 0.8615 | 0.8568 | +0.005 | D |
| CVV | FINANCIAL_ID | 362 | 0.8537 | 0.8360 | +0.018 | D |
| IBAN | FINANCIAL_ID | 769 | 0.8695 | 0.4952 | +0.374 | D |
| SWIFT_BIC | FINANCIAL_ID | 592 | 0.8423 | 0.8189 | +0.023 | D |
| SWIFT_BIC_CODE | FINANCIAL_ID | 122 | 0.6324 | 0.5172 | +0.115 | D |
| BANK_ROUTING_NUMBER | FINANCIAL_ID | 832 | 0.8214 | 0.8005 | +0.021 | D |
| ADDRESS | LOCATION | 6,490 | 0.6179 | 0.5269 | +0.091 | D |
| CITY | LOCATION | 2,036 | 0.8512 | 0.8324 | +0.019 | D |
| COORDINATE | LOCATION | 857 | 0.8874 | 0.9068 | −0.020 | S |
| COUNTRY | LOCATION | 2,887 | 0.8931 | 0.8642 | +0.029 | D |
| COUNTY | LOCATION | 611 | 0.8980 | 0.9275 | −0.030 | S |
| LOC | LOCATION | 32,846 | 0.6879 | 0.6441 | +0.044 | D |
| LOCAL_LATLNG | LOCATION | 55 | 0.3810 | 0.4655 | −0.085 | S |
| POSTCODE | LOCATION | 798 | 0.9021 | 0.8652 | +0.037 | D |
| STATE | LOCATION | 1,974 | 0.8291 | 0.8384 | −0.009 | S |
| STREET_ADDRESS | LOCATION | 4,557 | 0.7411 | 0.7154 | +0.026 | D |
| AGE | PERSON_GROUP | 454 | 0.8427 | 0.8447 | −0.002 | S |
| FIRST_NAME | PERSON_GROUP | 6,315 | 0.9326 | 0.9003 | +0.032 | D |
| GENDER | PERSON_GROUP | 509 | 0.9118 | 0.9082 | +0.004 | D |
| LAST_NAME | PERSON_GROUP | 4,684 | 0.9289 | 0.8777 | +0.051 | D |
| NAME | PERSON_GROUP | 6,645 | 0.5464 | 0.4355 | +0.111 | D |
| PERSON | PERSON_GROUP | 28,182 | 0.8066 | 0.7674 | +0.039 | D |
| COMPANY | ORG_ROLE | 4,039 | 0.5686 | 0.5688 | −0.000 | S |
| COMPANY_NAME | ORG_ROLE | 8,674 | 0.8487 | 0.8560 | −0.007 | S |
| JOB | ORG_ROLE | 2,645 | 0.9249 | 0.9129 | +0.012 | D |
| OCCUPATION | ORG_ROLE | 3,136 | 0.4726 | 0.4808 | −0.008 | S |
| ORG | ORG_ROLE | 12,229 | 0.7537 | 0.7061 | +0.048 | D |
| DATE | TEMPORAL | 18,080 | 0.6721 | 0.6215 | +0.051 | D |
| DATE_OF_BIRTH | TEMPORAL | 1,465 | 0.9497 | 0.9540 | −0.004 | S |
| DATE_TIME | TEMPORAL | 1,463 | 0.7422 | 0.7844 | −0.042 | S |
| TIME | TEMPORAL | 9,675 | 0.3665 | 0.3348 | +0.032 | D |
| DEVICE_IDENTIFIER | NETWORK | 496 | 0.8872 | 0.8774 | +0.010 | D |
| HTTP_COOKIE | NETWORK | 595 | 0.1812 | 0.5752 | −0.394 | S |
| IP_ADDRESS | NETWORK | 6,178 | 0.5528 | 0.4198 | +0.133 | D |
| IPV4 | NETWORK | 1,042 | 0.8395 | 0.8448 | −0.005 | S |
| IPV6 | NETWORK | 591 | 0.8421 | 0.8459 | −0.004 | S |
| MAC_ADDRESS | NETWORK | 606 | 0.8397 | 0.8412 | −0.002 | S |
| URL | NETWORK | 5,319 | 0.9003 | 0.8663 | +0.034 | D |
| USERNAME | NETWORK | 8,287 | 0.5285 | 0.4185 | +0.110 | D |
| USER_NAME | NETWORK | 1,292 | 0.8685 | 0.8598 | +0.009 | D |
| AMOUNT | MISC | 872 | 0.4033 | 0.2444 | +0.159 | D |
| BLOOD_TYPE | MISC | 462 | 0.8265 | 0.8862 | −0.060 | S |
| CC_SECURITY_CODE | MISC | 81 | 0.5920 | 0.2897 | +0.302 | D |
| CRYPTO_ADDRESS | MISC | 1,569 | 0.8641 | 0.0012 | +0.863 | D |
| CURRENCY | MISC | 1,931 | 0.8878 | 0.8290 | +0.059 | D |
| EDUCATION_LEVEL | MISC | 1,137 | 0.7623 | 0.7799 | −0.018 | S |
| EMPLOYMENT_STATUS | MISC | 720 | 0.8789 | 0.8571 | +0.022 | D |
| LANGUAGE | MISC | 669 | 0.8457 | 0.8209 | +0.025 | D |
| LICENSE_PLATE | MISC | 719 | 0.9444 | 0.9474 | −0.003 | S |
| MISC | MISC | 12,092 | 0.6334 | 0.5973 | +0.036 | D |
| POLITICAL_VIEW | MISC | 383 | 0.8370 | 0.8287 | +0.008 | D |
| RACE_ETHNICITY | MISC | 750 | 0.8541 | 0.8538 | +0.000 | D |
| RELIGIOUS_BELIEF | MISC | 452 | 0.8443 | 0.8502 | −0.006 | S |
| SEXUALITY | MISC | 230 | 0.9065 | 0.8993 | +0.007 | D |
| VEHICLE | MISC | 591 | 0.9787 | 0.4689 | +0.510 | D |
| VEHICLE_IDENTIFIER | MISC | 916 | 0.9467 | 0.9494 | −0.003 | S |

| EMAIL | CONTACT | 13,117 | 0.6373 | 0.5848 | +0.053 | D |
|---|---|---|---|---|---|---|
| FAX_NUMBER | CONTACT | 706 | 0.7504 | 0.7568 | −0.006 | S |
| PHONE | CONTACT | 1,535 | 0.9954 | 0.8213 | +0.174 | D |
| PHONE_NUMBER | CONTACT | 3,079 | 0.8606 | 0.8738 | −0.013 | S |
| ACCOUNT_PIN | CREDENTIAL | 91 | 0.5890 | 0.3333 | +0.256 | D |
| API_KEY | CREDENTIAL | 693 | 0.8167 | 0.8103 | +0.007 | D |
| BIOMETRIC_ID | CREDENTIAL | 1,090 | 0.9078 | 0.9082 | −0.000 | S |
| CERT_LICENSE_NUM | CREDENTIAL | 760 | 0.9122 | 0.9056 | +0.007 | D |
| CUSTOMER_ID | CREDENTIAL | 1,882 | 0.8876 | 0.8887 | −0.001 | S |
| DRIVER_LICENSE | CREDENTIAL | 80 | 0.6550 | 0.6447 | +0.010 | D |
| EMPLOYEE_ID | CREDENTIAL | 1,334 | 0.8669 | 0.8499 | +0.017 | D |
| HP_BENEF_NUMBER | CREDENTIAL | 640 | 0.8915 | 0.8941 | −0.003 | S |
| MEDICAL_RECORD | CREDENTIAL | 734 | 0.9658 | 0.9666 | −0.001 | S |
| NATIONAL_ID | CREDENTIAL | 300 | 0.8534 | 0.8561 | −0.003 | S |
| PASSPORT_NUMBER | CREDENTIAL | 115 | 0.6772 | 0.6190 | +0.058 | D |
| PASSWORD | CREDENTIAL | 2,510 | 0.9070 | 0.8655 | +0.042 | D |
| PIN | CREDENTIAL | 937 | 0.8686 | 0.8054 | +0.063 | D |
| SSN | CREDENTIAL | 1,261 | 0.9517 | 0.8261 | +0.126 | D |
| TAX_ID | CREDENTIAL | 252 | 0.8643 | 0.8683 | −0.004 | S |
| UNIQUE_ID | CREDENTIAL | 203 | 0.8064 | 0.8276 | −0.021 | S |
| FINANCIAL_ENTITY | FINANCIAL_NER | 58,821 | 0.3229 | 0.2412 | +0.082 | D |

*Table A1: Full entity-level comparison on the 100,002-record test split. W: D=Direct wins, S=SC+H wins.*